\documentclass[runningheads,a4paper]{llncs}

\usepackage{amssymb}
\usepackage{graphicx}
\usepackage{epstopdf}

\usepackage{url}
\urldef{\mailsa}\path|{amusleh, ndurrani, itemnikova, pnakov, svogel}@qf.org.qa|
\urldef{\mailsb}\path|osama.alsaad@qatar.tamu.edu|
\urldef{\mailsc}\path|erika.siebert-cole, peter.strasser, lncs}@springer.com|    
\newcommand{\keywords}[1]{\par\addvspace\baselineskip
\noindent\keywordname\enspace\ignorespaces#1}

\begin{document}
 
\mainmatter  

\title{Enabling Medical Translation\\
for Low-Resource Languages}

\titlerunning{Enabling Medical Translation for Low-Resource Languages}

%
%
\author{Ahmad Musleh$^{\dagger}$%
\and Nadir Durrani$^{\dagger}$ \and Irina Temnikova$^{\dagger}$ \and Preslav Nakov$^{\dagger}$\and\\
Stephan Vogel$^{\dagger}$ \and Osama Alsaad$^{\ddagger}$}
%

\institute{$^{\dagger}$Qatar Computing Research Institute, HBKU \hspace{5mm} $^{\ddagger}$Texas A\&M University in Qatar\\
\mailsa\\
\mailsb\\
}

%
%

\toctitle{Lecture Notes in Computer Science}
\tocauthor{Authors' Instructions}
\maketitle

\begin{abstract}

We present research towards bridging the language gap between migrant workers in Qatar and medical staff. In particular, we present the first steps towards the development of a real-world Hindi-English  machine translation system for doctor-patient communication. As this is a low-resource language pair, especially for speech and for the medical domain, our initial focus has been on gathering suitable training data from various sources. We applied a variety of methods  ranging from fully automatic extraction from the Web to manual annotation of test data. Moreover, we developed a 
method for automatically augmenting the training data with synthetically generated variants, which yielded a very sizable improvement of more than 3 BLEU points absolute.

\end{abstract}

\section{Introduction}
\label{introdction}
In recent years, Qatar's booming economy has resulted in rapid growth in the number of migrant workers needed for the growing number of infrastructure projects. These workers, who mainly come from the southern parts of Asia, usually know little or no English and do not know any Arabic either. This results in a communication barrier between them and the natives. More serious situation arises in the case of medical emergency.
This causes serious problems as the public administration and services in Qatar mostly use Arabic and English. 
According to a 2012 report by the Weill Cornell Medical College (WCMC\footnote{http://qatar-weill.cornell.edu}) in Qatar \cite{elnashar2012cultural}, almost 78\% of the patients visiting the Hamad Medical Corporation (HMC, the main health-care provider in Qatar) did not speak Arabic or English. The study also pointed out that the five most spoken languages in Qatar in 2012 were (in that order) Nepali, Urdu, Hindi, English, and Arabic. The report also pointed out that even though HMC currently uses medical interpreters to overcome this problem, their number is not sufficient. This has urged the authorities to look into technology for alternatives. 

In this paper, we propose a solution to bridge this language gap. We present our preliminary effort towards developing a Statistical Machine Translation (SMT) system for doctor-patient communication in Qatar. 
The success of a data-driven system largely depends upon the availability of in-domain data. 

This makes our task non-trivial as we are dealing with a low-resourced language pair and furthermore with the medical domain. We decided to focus on Hindi, one of the languages under question. Our decision was driven by the fact, that Hindi and Urdu are closely-related languages, often considered dialects of each other, and people from Nepal and other South-Asian countries working in Qatar typically understand Hindi. Moreover, we have access to more Hindi-English parallel data than for any other language pair involving the top-5 most spoken languages in Qatar.

Our focus in this paper is on data collection and data generation (Sections \ref{data_collection} and \ref{data_synthesis}).  We collected data from various sources (Section \ref{data_collection}), including (\emph{i})~Wikimedia (Wikipedia, Wiktionary, and OmegaWiki) parallel English-Hindi data, (\emph{ii})~doctor-patient dialogues from YouTube videos, and movie subtitles, and (\emph{iii})~parallel medical terms from BabelNet and MeSH.
Moreover, we synthesized Hindi-English parallel data from Urdu-English data, by translating the Urdu part into Hindi. The approach is described in Section \ref{data_synthesis}. 
Our results show improvement of up to +1.45 when using synthesized data, and up to +1.66, when concatenating the mined dictionaries on top of the synthesized data (Section \ref{experiments_and_evaluation}).

Moreover, Section \ref{related_work} provides an overview of related work on machine translation for doctor-patient dialogues and briefly discusses the Machine Translation (MT) approaches for low-resource languages; 
Section~\ref{experiments_and_evaluation} presents the results of the manual evaluation, and Section \ref{conclusions} provides the conclusions and discusses directions for future work.

\section{Related Work}
\label{related_work}Below we first describe some MT applications for doctor-patient communication. Then, we present more general research on MT for the medical domain.

\subsection{Bi-directional Doctor$\leftrightarrow$Patient Communication Applications}

We will first describe the pre-existing MT systems for doctor-patient communication, particularly the ones that required data collection for under-resourced languages. Several MT systems facilitating doctor-patient communication have been built in the past \cite{bouillon2008many,eck2010jibbigo,dillinger2006converser,ehsani2006speech,heinze2006automated,gao2006ibm}. 
Most of them are still prototypes, and only few have been fully deployed.
Some of these systems work with under-resourced languages \cite{bouillon2008many,ehsani2006speech,heinze2006automated,gao2006ibm}. Moreover, their solution 
relies on mapping the utterances to an interlingua, instead of using SMT.

\textbf{MedSLT} 
\cite{bouillon2008many} is an interlingua-based speech-to-speech translation system. It covers a restricted set of domains, and covers English, French, Japanese, Spanish, Catalan, and Arabic. The system can translate doctor's questions/statements to the patient, but not the responses by the patient back to the doctor.

\textbf{Converser}  \cite{dillinger2006converser} is a commercial doctor-patient speech-to-speech bidirectional MT system for the English$\leftrightarrow$Spanish language pair. It has been deployed in several 
US hospitals and has the following features: users can correct the automatic speech recognition (ASR) and MT outputs, back-translation (re-translation of the translation) to the user is made to allow this. The system maps concepts to a lexical database specially created from various sources, and also allows ``translation shortcuts'' (i.e., translation memory of previous translations that do not need verification).

\textbf{Jibbigo} \cite{eck2010jibbigo} is a travel and medical speech-to-speech MT system, deployed as an iPhone mobile app.  Jibbigo allows English-to-Spanish and Spanish-to-English medical speech-to-speech translation.

\textbf{S-MINDS} \cite{ehsani2006speech} is a two-way doctors-patient MT system, which 
uses an in-house ``interpretation'' software. 
It matches the ASR utterances to a finite set of concepts in the specific semantic domain and then paraphrases them. In case the utterance cannot be matched, the system uses an SMT engine.

\textbf{Accultran}  \cite{heinze2006automated}
is 
an automatic translation system prototype, which features back translation to the doctor and yes/no or multiple-choice questions (MCQs) to the patient. It allows the doctor to confirm the translation to the patient, and has a cross cultural adviser. It flags sensitive utterances that are difficult to translate to the patient. The system maps the utterances to SNOMED-CT or Clinical Document Architecture (CDA-2) standards, which are used as an interlingua. 

\textbf{IBM MASTOR} \cite{gao2006ibm}
is a speech-to-speech MT system for two language pairs (English-Mandarin and English-dialectal Arabic), which relies on ASR, SMT, and Speech Synthesis components. It works both on laptops and PDAs. 

\textbf{English-Portugese SLT} \cite{rodrigues2013speech} is an English-Portuguese speech-to-speech MT system, composed of an ASR, MT (relying on HMM) and speech synthesis. It works as an online service and as a mobile application. 

None of the above systems handles the top-5 languages of interest to Qatar. 

\subsection{Uni-directional Doctor$\leftrightarrow$Patient Communication Applications}

Besides the above-described MT systems, there are a number of mobile or web applications, which are based on pre-translated phrases. The phrases are pre-recorded by professionals or native speakers and can be played to the patient. Most of these applications work only in the doctor-to-patient direction. The most popular ones are UniversalDoctor\footnote{http://www.universaldoctor.com}, MediBabble\footnote{http://medibabble.com}, Canopy\footnote{http://www.canopyapps.com}, MedSpeak, Mavro Emergency Medical Spanish\footnote{http://mavroinc.com/medical.html}, and DuoChart\footnote{http://duochart.com}. 

Unfortunately, these applications do not allow free, unseen, or spontaneous translations, and do not cover the language pairs of interest for Qatar. Moreover, some of them (e.g., UniversalDoctor) require a paid subscription.

\subsection{General Research in Medical Machine Translation}

A number of systems have been developed and participated in the WMT'14 Medical translation task. It is a Cross-Lingual Information Retrieval (CLIR) task divided into two sub-tasks: (\emph{i}) translation of user search queries, and (\emph{ii})~translation of summaries of retrieved documents:

A system described in \cite{duvsek2014machine}, part of the Khresmoi project\footnote{http://www.khresmoi.eu}, uses the phrase-based Moses  and standard methods for domain adaptation.
\cite{li2014postech} also uses the phrase-based Moses system and achieved the highest BLEU score for the English-German intrinsic query translation evaluation. Another system \cite{lu2014domain} combined web-crawled in-domain monolingual data and a bilingual lexicon in order to complement the limited in-domain parallel corpora. A third one \cite{okita2014dcu} proposed a terminology translation system for the query translation subtask and used 6 different methods for terminology extraction. A fourth system \cite{pecheux2014limsi} used a combination of n-gram based NCODE and phrase-based Moses to the subtask of sentence translation. The system of \cite{wang2014combining} applied a combination of domain adaptation techniques on the medical summary sentence translation task and achieved the first and the second best BLEU scores. Then, the system of \cite{zhang2014experiments} used the Moses phrase-based system and  worked on the medical summary WMT'14 task and experimented with translation models, re-ordering models, operation sequence models, and language models, as well as with data selection. A study on quality analysis of machine translation systems in medical domain was carried in \cite{martaMedical}. Most of this work focused on European language pairs and did not cover languages of interest to us, nor did it involve low-resource languages in general.

\section{Data Collection}
\label{data_collection}As the main problem of low resource languages is data collection \cite{lewis2011crisis,nakov2012improving}, we have adopted a variety of approaches, in order to collect as much parallel English-Hindi data as possible.



\subsection{Wiki Dumps}
 
We downloaded, extracted and mined all language links from Wikipedia,\footnote{http://www.wikipedia.org} Wiktionary,\footnote{http://www.wiktionary.org} and OmegaWiki\footnote{http://www.omegawiki.org} in order to provide a one-to-one word mapping from English into Hindi.
We then extracted page links and language links from Wikipedia and Wiktionary.
Moreover, we used OmegaWiki to provide a bilingual word dictionary containing the word, its synonyms, its translation, and its
lexical, terminological and ontological forms. We extracted the data using two OmegaWiki sources: bilingual dictionaries and an SQL database dump. Table~\ref{tab:wiki} shows the number of Hindi words we collected from all three sources. 

\begin{table}
\begin{center}
\begin{tabular}{|c|c|}
\hline
\bf Source & \bf Word Pairs\\
\hline
Wikipedia & 40,764\\
Wiktionary & 10,352\\
OmegaWiki & 3,476\\
\hline
\end{tabular}
\end{center}
\caption{Number of word translation pairs collected from Wikimedia sources.}
\label{tab:wiki}
\end{table}

\subsection{Doctor-Patient YouTube Videos and Movie Subtitles}

We used a mixed approach to extract doctor-patient dialogues from medical YouTube subtitles. As using the YouTube-embedded automatic subtitling is inefficient, the dialogues of the videos were first extracted by manually typing the audio found in the videos. However, as this process was very time consuming, we started a screenshot session in order to collect all the visual representations of the subtitles. Next, the subtitles were extracted using Tesseract,\footnote{https://github.com/tesseract-ocr} an open source Optical Character Recognition (OCR) reader provided by Google, on the screenshots captured. The subtitles were then manually corrected, translated into Hindi using Google Translate, and post-edited by a Hindi native speaker. This resulted in a parallel corpus of medical dialogues with 11,000 Hindi words (1,200 sentences). These sentences were later used for tuning and testing our MT system. Additionally, we used a web crawler to extract a small number of non-medical parallel English-Hindi movie subtitles (nine movies) from Open Subtitles.\footnote{http://www.opensubtitles.com}



\subsection{BabelNet and MeSH}

We extracted medical terms from BabelNet \cite{NavigliPonzetto:12aij} using their API. As Medical Subject Headings (MeSH\footnote{http://www.ncbi.nlm.nih.gov/mesh}) represents the largest source of Medical terms, we downloaded their dumps and extracted the 198,958 MeSH terms which we overlapped with the previously mined results of Wiki Dumps.

\section{Data Synthesis}
\label{data_synthesis}Hindi and Urdu are closely-related languages that share grammatical structure and largely overlap in vocabulary. This provides strong motivation to transform an Urdu-English parallel data into Hindi-English by translating the Urdu part into Hindi. 
We made use of the Urdu-English segment of the Indic multi-parallel corpus \cite{post-callisonburch-osborne:2012:WMT}, which contains about 87,000 sentence pairs. The Hindi-English segment of this corpus is a subset of the parallel data that was made available for the WMT'14 translation task, but its English side is completely disjoint from the English side of the Urdu-English segment. 


Initially, we trained an Urdu-to-Hindi SMT system using the tiny EMILLE\footnote{EMILLE contains about 12,000 sentences of comparable data in Hindi and Urdu. We were able to align about 7,000 sentences to build an Urdu-to-Hindi system.} corpus \cite{BakerHMCG02}. However, we found this system to be useless for translating the Urdu part of the Indic data due to domain mismatch and the high proportion of Out-of-Vocabulary (OOV) words (approximately 310,000 tokens). Thus, in order to reduce data sparseness, we synthesized additional phrase tables using interpolation and transliteration.

\subsection{Interpolation}

We built two phrase translation tables $p(\bar{u_i}|\bar{e_i})$ and $p(\bar{e_i}|\bar{h_i})$, from Urdu-English (Indic corpus) and Hindi-English (HindEnCorp \cite{hindencorp:pbml:2014}) bitexts. Given the phrase table for Urdu-English $p(\bar{u_i}|\bar{e_i})$ and the phrase table for English-Hindi $p(\bar{e_i}|\bar{h_i})$, we induced an Urdu-Hindi phrase table $p(\bar{u_i}|\bar{h_i})$ using the model \cite{UtiyamaI07,wu-wang:2007:ACLMain}:

\begin{equation}
p(\bar{u_i}|\bar{h_i}) = \sum_{\bar{e_i}} p(\bar{u_i}|\bar{e_i})p(\bar{e_i}|\bar{h_i}) 
\nonumber
\end{equation}

\noindent The number of entries in the baseline Urdu-to-Hindi phrase table were approximately 254,000. Using interpolation, we were able to build a phrase table containing roughly 10M phrases. This reduced the number of OOV tokens from 310K to approximately 50,000.

\subsection{Transliteration}

As Urdu and Hindi are written in different scripts (Arabic and Devanagri, respectively), we added a transliteration component to our Urdu-to-Hindi system. While it can be used to translate all 50,000 OOV words, previous research has shown that transliteration is useful for more than just translating OOV words when translating closely related language pairs \cite{Nakov:2009:ISM,nakov-tiedemann:2012:ACL2012short,tiedemann-nakov:2013:RANLP-2013,wang-nakov-ng:2012:EMNLP-CoNLL,Pidong:al:CL:2016}. Following \cite{durrani-EtAl:2010:ACL}, we transliterate all Urdu words to Hindi and hypothesize n-best transliterations, along with regular translations. The idea is to generate novel Hindi translations that may be absent from the regular and interpolated phrase table, but for which there is evidence in the language model. Moreover, the overlapping evidence in the translation and transliteration phrase tables improves the overall system.

We learn an unsupervised transliteration model \cite{durrani-EtAl:2014:EACL} from the word-alignments of Urdu-Hindi parallel data. We were able to extract around 2,800 transliteration pairs.
To learn a richer transliteration model, we additionally fed the interpolated phrase table, as described above, to the transliteration miner. We were able to mine about 21,000 additional transliteration pairs and to build an Urdu-Hindi character-based model from it. In order to fully capitalize on the large overlap in Hindi--Urdu vocabulary, we transliterated each word in the Urdu test data to Hindi and we produced a phrase table with 100-best transliterations. We then used the two synthesized (triangulated and transliterated) phrase tables along with the baseline Urdu-to-Hindi phrase table in a log-linear model. 

Table \ref{tab:ur-hi} shows development results from training an Urdu-to-Hindi SMT system. By adding interpolated phrase tables and transliteration, we obtain a very sizable gain of +3.35 over the baseline Urdu$\rightarrow$Hindi system. Using our best Urdu-to-Hindi system (${\bf B_{u,h}}{\bf T_g}{\bf T_r}$), we translated the Urdu part of the multi-Indic corpus to form a Hindi-English bi-text. This yielded a synthesized bi-text of $\approx$87,000 Hindi-English sentence pairs. Detailed analysis can be found in \cite{Durrani-pivot-eamt14}.

\begin{table}[htb]
\centering
\begin{tabular}{|l|c|c|c||l|c|c||c|c|c|}
\hline
{\bf System}	& {\bf PT} & {\bf Tune} & {\bf Test}  & {\bf System} & {\bf Tune} & {\bf Test} & {\bf System} & {\bf Tune} & {\bf Test} \\
\hline
${\bf B_{u,h}}$ & 254K & 34.18 & 34.79 & ${\bf B_{u,h}}{\bf T_g}$  & 37.65 & 37.58 & & & \\
${\bf T_g}$  &  10M & 15.60 &  15.34 & ${\bf B_{u,h}}{\bf T_r}$ & 34.77 & 35.76 & ${\bf B_{u,h}}{\bf T_g}{\bf T_r}$ & 38.0 & 37.99\\
${\bf T_r}$ &  & 9.54 &  9.93 & ${\bf T_g}{\bf T_r}$  & 17.63 & 18.11 &  & $\Delta$+3.89 & $\Delta$+3.35 \\
\hline
\end{tabular}
\vspace{2mm}
\caption{Evaluating triangulated and transliterated phrase tables for Urdu-to-Hindi SMT. Notation: ${\bf B_{u,h}}$ = Baseline Phrase Table, ${\bf T_g}$ = Triangulated Phrase Table, and ${\bf T_r}$ = Transliteration Phrase Table.}
\label{tab:ur-hi}
\end{table}

\section{Experiments and Evaluation}
\label{experiments_and_evaluation}\subsection{Machine Translation}
\subsubsection{Baseline Data}

We trained the Hindi-English systems using Hindi-English parallel data \cite{hindencorp:pbml:2014} composed by compiling several sources including the Hindi-English segment of the Indic parallel corpus. It contains 287,202 parallel sentences, and 4,296,007 Hindi and 4,026,860 English tokens. We used 635 sentences (6,111 tokens) for tuning and 636 sentences (5,231) for testing, collected from doctor-patient communication dialogues in YouTube videos. The sentences were translated into Hindi by a human translator. We trained interpolated language models using all the English and Hindi monolingual data made available for the WMT'14 translation task: 287.3M English and 43.4M Hindi tokens.

\subsubsection{Baseline System}

We trained a phrase-based system using Moses \cite{koehn07:moses} with the following settings: a maximum sentence length of 80, GDFA symmetrization of GIZA++ alignments  \cite{och03:asc}, an interpolated Kneser-Ney smoothed 5-gram language model with KenLM \cite{Heafield-kenlm} used at runtime, 100-best translation options, MBR decoding \cite{KumarB04}, Cube Pruning \cite{huang-chiang:2007:ACLMain} using a stack size of 1,000 during tuning and 5,000 during testing. We tuned with the $k$-best batch MIRA \cite{cherry-foster:2012:NAACL-HLT}. We additionally used msd-bidirectional-fe lexicalized reordering, a 5-gram OSM \cite{durrani15:cl}, class-based models \cite{Durrani-osm-coling14}\footnote{We used mkcls to cluster the data into 50 clusters.} sparse lexical and domain features \cite{iwslt12:Hasler-2}, a distortion limit of 6, and the no-reordering-over-punctuation heuristic. We used an unsupervised transliteration model \cite{durrani-EtAl:2014:EACL} to transliterate the OOV words. These are state-of-the-art settings, as used in \cite{Durrani-uedinwmt14}.

\subsubsection{Results}

\begin{table}[h]
\begin{center}
\begin{tabular}{|l|c|c|c|c|c|}
\hline
Pair & ${\bf B_0}$ & ${\bf +Syn}$ & $\Delta$ & ${\bf +PT}$ & $\Delta$ \\
\hline
hi-en & 21.28 & 22.67 & +1.39 & 22.1 & +0.82 \\
en-hi & 22.52 & 23.97 & +1.45 & 23.28 & +0.76 \\
\hline
\end{tabular}
\end{center}
\caption{Results using synthesized (Syn) Hindi-English parallel data. Notation used: ${\bf B_0}$ = System without synthesized data, ${\bf +PT}$ = System using synthesized data as an additional phrase table.}
\label{tab:syn}
\end{table}

Table \ref{tab:syn} shows the results when adding the synthesized Hindi-English bi-text on top of the baseline system (+Syn).  The synthesized data was simply concatenated to the baseline data to train the system. We also tried building phrase tables (+PT) separately from the baseline data and from the synthesized one and used as a separate features in the log-linear model as done in \cite{nakov:2008:WMT,Nakov:2009:ISM,wang-nakov-ng:2012:EMNLP-CoNLL,Pidong:al:CL:2016}. We found that concatenating synthetic data with the baseline data directly was superior to training a separate phrase-table from it. We obtained improvements of up to +1.45 by adding synthetic data.

Table \ref{tab:dict} shows the results from adding the mined dictionaries to the baseline system. The baseline system (${\bf B_0}$) used in this case is the best system in Table~\ref{tab:syn}. Again, we simply concatenated the dictionaries with the baseline data and we gained improvements of up to +1.66 BLEU points absolute. Cumulatively, by using dictionaries and synthesized phrase-tables, we were able to obtain statistically significant improvements of more than 3 BLEU points.

\begin{table}[h]
\begin{center}
\begin{tabular}{|l|c|c|c|}
\hline
Pair & ${\bf B_0}$ & ${\bf +Dict}$ & $\Delta$ \\
\hline
hi-en & 22.67 & 23.29 & +0.62 \\
en-hi & 23.97 & 25.63 & +1.66 \\
\hline
\end{tabular}
\end{center}
\caption{Evaluating the effect of Dictionaries. ${\bf B_0}$ = System without dictionaries.}
\label{tab:dict}
\end{table}

\subsection{Manual Evaluation}
In addition to the above evaluation, we ran a small manual evaluation experiment, using the Appraise platform \cite{federmann2012appraise}. The two sections below describe the results of the Hindi-to-English (Section \ref{hindi_to_english}) and the English-to-Hindi (Section \ref{english_to_hindi}) evaluations. 

\subsubsection{Hindi-to-English} \label{hindi_to_english}
The evaluation was conducted by 3 monolingual English speakers, using 321 randomly selected sentences, divided into three batches of evaluation. 
Similar to the setup at evaluation campaigns such as WMT,
the evaluators were shown the 
translations and references. 

The evaluators were asked to assign one of the following three categories to each translation: (a)~\emph{helpful in this situation}, (b)~\emph{misleading}, and (c)~\emph{doubtful that people will understand it}. As shown in Table \ref{tab:hi-en-Man}, over 37\% of the cases were classified as \emph{helpful} (good translations), 39\% as doubtful (mediocre), and 24\% as \emph{misleading} (really bad translations). Annotators did not always agree, e.g.,~Judge 1 and Judge 2 were more lenient than Judge 3. 

\begin{table}\label{tab:hin_en}
\begin{center}
\begin{tabular}{|c|c|c|c|c|c|}
\hline
\bf Response & \bf Judge 1 & \bf Judge 2 & \bf Judge 3 & \bf Total & \bf Percentage \\
\hline
Helpful & 129 & 129 & 96 & 354 & 37\%\\
Doubtful & 114 & 145 & 118 & 377 & 39\%\\
Misleading & 78 & 47 & 107 & 232 & 24\%\\
\hline
\end{tabular}
\end{center}
\caption{Manual sentence evaluation for Hindi-to-English translation.}
\label{tab:hi-en-Man}
\end{table}

\subsubsection{English-to-Hindi}\label{english_to_hindi}

In order to check the output of the English-to-Hindi system, we asked a bilingual judge to evaluate 328 sentences. She was asked to classify the sentences in the same categories as for the Hindi-English evaluation. Table \ref{tab:en-hi-Man} shows the results; we can see that 55.8\% of the sentences were found \emph{helpful in this situation}. 
This is hardly because English-Hindi system was any better, but more likely because the human evaluator was lenient. Unfortunately, we could not find a second Hindi speaker to evaluate our translations, and thus we could not calculate inter-annotator agreement.

\begin{table}\label{tab:en_hin}
\begin{center}
\begin{tabular}{|c|c|c|}
\hline
\bf Response & \bf Number & \bf Percentage\\
\hline
Helpful & 183 & 55.8\% \\
Doubtful & 111 & 33.8\% \\
Misleading & 34 & 10.4\%\\
\hline
\end{tabular}
\end{center}
\caption{Manual sentence evaluation for English-to-Hindi translation.}
\label{tab:en-hi-Man}
\end{table}

\subsubsection{Analysis} 
In order to understand the problems with the Hindi output, we conducted  an error analysis on 100 sentences classifying the errors into the following categories:

\begin{itemize}
\item missing/untranslated words;
\item wrongly translated words; 
\item word order problems; 
\item other error types, e.g., extra words.
\end{itemize}

Table \ref{tab:errors} shows the results. We can see that most of the problems are associated with \emph{word order problems} (84\%) or \emph{wrongly translated words} (74\%). 

\begin{table} \label{tab:error_analysis}
\begin{center}
\begin{tabular}{|c|c|}
\hline
\bf Category & \bf Percentage \\
\hline
Missing/Untranslated words & 45\%\\
Wrongly translated words & 74\%\\
Word order problems & 84\%\\
Other types of errors & 13\%\\
\hline
\end{tabular}
\end{center}
\caption{Error analysis results for our English-to-Hindi translation.}
\label{tab:errors}
\end{table}

\section{Conclusions}
\label{conclusions}We presented our preliminary efforts towards building a Hindi$\leftrightarrow$English SMT system for facilitating doctor-patient communication. We improved our baseline system using two approaches, namely (\emph{i})~additional data collection, and (\emph{ii})~automatic data synthesis. We mined useful dictionaries from Wikipedia in order to improve the coverage of our system. We made use of the relatedness between Hindi and Urdu to generate synthetic Hindi-English bi-texts by automatically translating 87,000 Urdu sentences into Hindi. Both our data collection and our synthesis approach worked well and have shown significant improvements over the baseline system, yielding a total improvement of +3.11 BLEU points absolute for English-to-Hindi and +2.07 for Hindi-to-English. We also carried out human evaluation for the best system. 
In the error analysis of the Hindi outputs, we found that most errors were due to ordering of the words in the output, or to wrong lexical choice.

In future work, we plan to collect more data for Hindi, but also to synthesize Urdu data. We further plan to develop a system for Nepali-English. Finally, we would like to add Automatic Speech Recognition (ASR) and Speech Synthesis components in order to build a fully-functional speech-to-speech system, which we would test and gradually deploy for use in real-world scenarios.

\section*{Acknowledgments}
The authors would like to thank Naila Khalisha and Manisha Bansal for their contributions towards the project.

\bibliography{MTFDPC.bib}
\bibliographystyle{plain}

\end{document}